\pdfoutput=1
\documentclass[11pt]{article}

\usepackage[final]{acl}

\usepackage{times}
\usepackage{latexsym}

\usepackage[T1]{fontenc}
\usepackage[utf8]{inputenc}

\usepackage{microtype}

\usepackage{inconsolata}
\usepackage{pdfpages}

\usepackage{graphicx}
\usepackage{amsmath}
\usepackage{amsfonts}
\usepackage{booktabs, multirow} 
\usepackage{adjustbox}
\usepackage{arydshln}
\usepackage{array}
\title{TECO: Improving  Multimodal Intent Recognition with Text Enhancement through Commonsense Knowledge Extraction}

\author{Quynh-Mai Thi Nguyen, Lan-Nhi Thi Nguyen, Cam-Van Thi Nguyen\thanks{Corresponding author.
	Cam-Van Thi Nguyen was funded by the Master, PhD Scholarship Programme of Vingroup Innovation Foundation (VINIF), code VINIF.2023.TS147.} \\
        Faculty of Information Technology \\
        VNU University of Engineering and Technology \\
        \texttt{\{21020125, 21020372, vanntc\}@vnu.edu.vn}}

\begin{document}
\maketitle
\begin{abstract}
The objective of multimodal intent recognition (MIR) is to leverage various modalities—such as text, video, and audio—to detect user intentions, which is crucial for understanding human language and context in dialogue systems. Despite advances in this field, two main challenges persist: \textit{(1) effectively extracting and utilizing semantic information from robust textual features; (2) aligning and fusing non-verbal modalities with verbal ones effectively}. This paper proposes a \textbf{\underline{T}}ext \textbf{\underline{E}}nhancement with \textbf{\underline{C}}omm\textbf{\underline{O}}nsense Knowledge Extractor (TECO) to address these challenges. We begin by extracting relations from both generated and retrieved knowledge to enrich the contextual information in the text modality. Subsequently, we align and integrate visual and acoustic representations with these enhanced text features to form a cohesive multimodal representation. Our experimental results show substantial improvements over existing baseline methods.
\end{abstract}

\section{Introduction}
Intent recognition plays a vital role in natural language understanding. While prior attempts focused on a single modality, e.g., text, for extraction \cite{hu2021semi}, real-world scenarios involve intricate human intentions that require the integration of information from speech, tone, expression, and action. Recently, multimodal intent recognition (MIR) performed computationally is a very interesting and challenging task to be explored. To effectively leverage the information from various modalities, numerous methods have been proposed for MIR. As an alternative, \cite{tsai2019multimodal}; \cite{rahman2020integrating} proposed frameworks using transformer-based techniques to integrate information from different modalities into a unified feature. Moreover, \cite{zhou2024token} introduced a token-level contrastive learning method with a modality-aware prompting module; \cite{huang2024sdif} proposed a shallow-to-deep transformer-based framework with ChatGPT-based data augmentation strategy, achieving an impressive result. Despite the advances, we suppose that existing MIR models still suffer from several challenges: (1) how to explore the semantic information from the contextual features effectively; (2) the limitation in aligning and fusing features of different modalities.
\begin{figure}[t!]
    \centering
    \includegraphics[width=\linewidth]{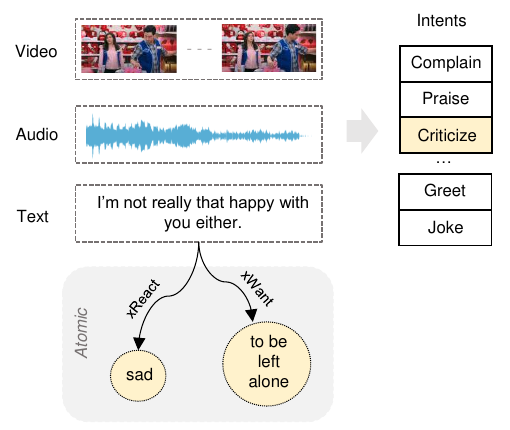}
    \caption{An example of integrating commonsense knowledge for multi-intent recognition provides awareness about implicit context which relates to the utterance's intention.}
    \label{fig:example}
\end{figure}

To address the above challenges, we introduce a framework called Text Enhancement with Commonsense Knowledge Extractor (\textbf{TECO}). Our model comprises three main components: a Commonsense Knowledge Extractor (COKE), a Textual Enhancement Module (TEM), and a Multimodal Alignment Fusion (MAF). Our main idea is to explore rich and comprehensive contextual features and then incorporate them with non-verbal features (image, audio) to predict the reasonable utterance of the participants. COKE combines both retrieved and generated commonsense knowledge to capture relational features, whereas TEM utilizes a dual perspective learning module and a textual enhancing fusion to integrate them into the text feature. Finally, we adopt MAF to effectively fuse features from three modalities into multimodal knowledge-enhanced representations of utterances. 

Our contributions are summarized as follows:
\begin{itemize}
    \item We propose the TECO model, featuring a Text Enhancement Module (TEM) with commonsense knowledge extraction to effectively leverage semantic information from textual input.
    \item TECO incorporates Dual Perspective Learning to integrate and harmonize relation perspectives and aligns non-verbal modalities with verbal ones for consistent multimodal representation.
    \item Experimental results and detailed analyses on the challenging MIntRec dataset demonstrates the superior performance of our TECO model in multimodal intent detection.
\end{itemize}

\section{Related Works}
\subsection{Commonsense Knowledge}
Commonsense reasoning utilizes the basic knowledge that reflects our natural understanding of the world and human behavior, which is crucial for interpreting the latent variables of a conversation. Recently, COMET \cite{bosselut2019comet} has achieved impressive performance when investigating and transferring implicit knowledge from a deep pre-trained language model to generate explicit knowledge in commonsense knowledge graphs. The seminal works utilize COMET to guide the participants through their reasoning about the content of the conversation, dialog planning, making decisions, and many reasoning tasks. SHARK \cite{wang2023shark} uses a pre-trained neural knowledge model COMET-ATOMIC \cite{Hwang2021COMETATOMIC2O} to extract emotion utterance by generating novel commonsense knowledge tuples, CSDGCN \cite{yu2023commonsense} proposed using COMET to clearly depict how external commonsense knowledge expressions within the context contributes to sarcasm detection, $R^3$ \cite{chakrabarty2020r} retrieve relevant context for the sarcastic messages based on commonsense knowledge. 

Sentence-BERT \cite{reimers2019sentence} uses siamese and triplet network structure to capture semantically meaningful sentence features that can compared using cosine-similarity. In this paper, we incorporate two views from generative and retrieved relations to enrich context information via two pre-trained models, COMET and SBERT.

\subsection{Multimodal Fusion}
Multimodal Fusion is an active area of research with various proposed methods. Prior studies based on transformer, MULT \cite{tsai2019multimodal} directly attend to elements in other modalities and capture long-range crossmodal events. However, it does not handle modality non-alignment by simply aligning them. Moreover, MAG-BERT \cite{rahman2020integrating}  proposed an efficient framework for fine-tuning BERT \cite{devlin2018bert} and XLNet \cite{yang2019xlnet} for multimodal input and MISA \cite{hazarika2020misa} projects each modality to two distinct subspaces, which provide a holistic view of the multimodal data. To effectively fuse different modalities's features and alleviate the data scarcity problem, SDIF-DA \cite{huang2024sdif} introduced a shallow-to-deep interaction framework using a hierarchical and a transformer module. Recent researches attempt to extract more information from textual input, Promt Me Up \cite{hu2023prompt} proposed innovative pre-training objects for entity-object and relation-image alignment, extracting objects from images and aligning them with entity and relation prompts. To leverage the limitations in learning semantic features, TCL-MAP \cite{zhou2024token} develops a token-level contrastive learning method with a modality-aware prompting module.

\begin{figure*}
    \centering
    \includegraphics[width=1\linewidth]{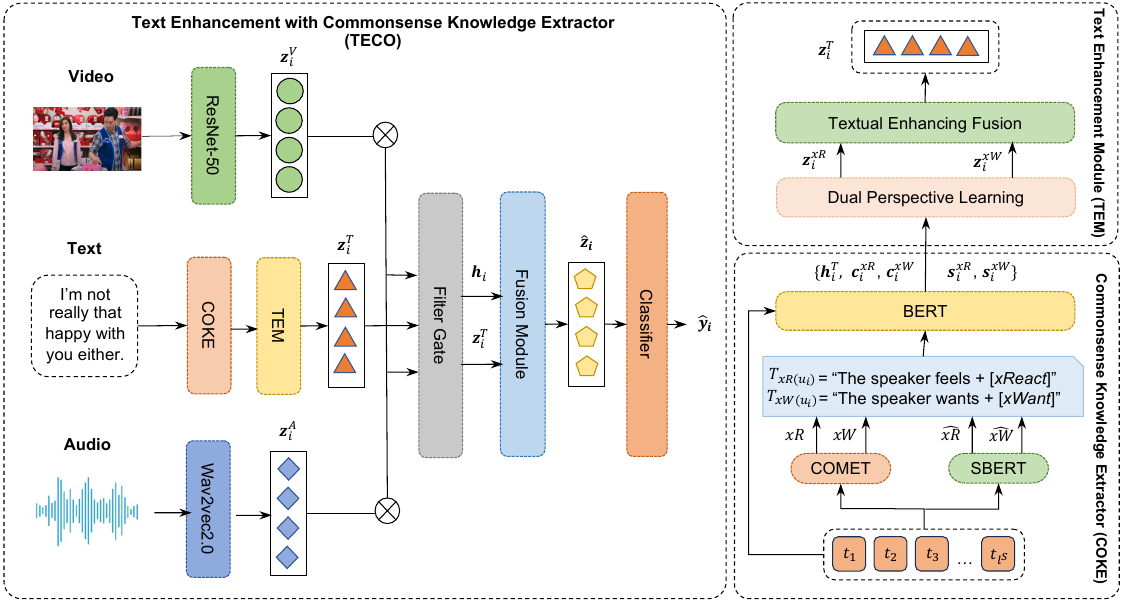}.
    \caption{Overall architecture of our model is illustrated in the left part. The lower right part describes the flow of the Commonsense Knowledge Extractor (COKE), and the upper one shows details of the Text Enhancement Module (TEM), which integrates relation features into textual representations using commonsense knowledge and a dual perspective learning module.}
    \label{fig:architecture}
\end{figure*}

\section{Methodology}
\subsection{Problem Statement and Model Overview}
\textbf{Problem Statement.}
Multi-modal intent recognition aims to analyze various modalities such as expression, body movement, and tone of speech to understand a user's intent. Given an input text $T = \{t_1, t_2, ..., t_{l^S}\}$ with the corresponding image $V$ and audio $A$, where $l^S$ is the length of the text sequence, our model is supposed to classify given text into correct intent category $i \in \mathbb{I} = \{i_1, i_2,..., i_N\}$. The set $\mathbb{I}$ contains the pre-defined intent types, and $N$ represents the number of utterances.

\textbf{Model Overview.}
Figure \ref{fig:architecture} describes the architecture of our model, which comprises three components. The input sentence is converted into vector representations using an encoding context module. Next, in the Textual Enhancement Module (TEM), we utilize a commonsense reasoning module to extract relevant knowledge and convert it into vector representations. Subsequently, the output vector is put into a dual mechanism to obtain a single representation. 

We also extract features from audio segments and video segments by using encoder mechanisms. After each extracted feature is aligned with the textual information, we concatenate the textual feature with the visual and acoustic information and utilize them to compute two filter gates, which emphasize relevant information from visual and acoustic modalities based on the textual input. We then separately feed each obtained feature into a fusion module. Finally, in the prediction stage, we perform a classifier operation to get the final utterance detection result.

\subsection{Feature Encoders}
For each utterance $u_i$, we extract multimodal features from three different modalities: text, vision, and audio.

\textbf{Textual Encoder.} The pre-trained BERT language model \cite{devlin2018bert} which achieves excellent performance in Natural Language Processing (NLP) is applied to extract text features. For each input sentence $t_i$, we obtain the token embeddings from the last hidden layer of the BERT Encoder:
\begin{equation}
    \textbf{h}_{i}^{T} = \textrm{TextEncoder}(t_i)
\end{equation}
where TextEncoder is BERT Encoder, $\textbf{h}_{i}^{T} \in \mathbb{R}^{l^S \times d}$ refers to the text embedding of text sentence $t_i$, $l^S$ is the length of text sentence, and $d$ denotes the feature dimension.

\textbf{Visual Encoder.} We follow the approach used in previous work \cite{zhang2022mintrec} to process video segments. By leveraging a pre-trained Faster R-CNN \cite{ren2015faster} with the backbone ResNet-50 \cite{koonce2021resnet}, the vision
feature embeddings are extracted as follows:
\begin{equation}
    \textbf{h}_{i}^{V} = \textrm{VisualEncoder}(v_i)
\end{equation}
where VisualEncoder is Faster R-CNN, $\textbf{h}_{i}^{V} \in \mathbb{R}^{l^V \times d^V}$ denotes the vision embedding of video segment $v_i$, $l^V$ is the length of video segment, and $d^V$ refers to the vision feature dimension.

\textbf{Acoustic Encoder.} To extract the acoustic embeddings, we utilize a pre-trained model wav2vec 2.0 \cite{schneider2019wav2vec}, which employs self-supervised learning to generate strong representations for speech recognition. The formula is shown as follows:
\begin{equation}
    \textbf{h}_{i}^{A} = \textrm{AcousticEncoder}(a_i)
\end{equation}
where AcousticEncoder refers to wav2vec 2.0, $\textbf{h}_{i}^{A} \in \mathbb{R}^{l^A \times d^A}$ denotes the acoustic embedding of audio segment $a_i$, $l^A$ is the audio segment's length, and $d^A$ denotes the acoustic feature dimension.

\subsection{Commonsense Knowledge Extractor (COKE)}
For each utterance, we utilize a commonsense knowledge graph combined with two pre-trained models to obtain relational features. Subsequently, integrating them into textual features to enhance textual information.

\textbf{Relation Generation.} We put each utterance through a pre-trained generative model COMET\footnote{\url{https://github.com/atcbosselut/comet-commonsense}} \cite{bosselut2019comet}, which is able to produce rich and diverse commonsense knowledge relying on a seed set of knowledge tuples. A knowledge base ATOMIC\footnote{\url{https://github.com/allenai/comet-atomic-2020/}} \cite{Hwang2021COMETATOMIC2O} is used as a knowledge seed set to generate phrases of several relation types. Among nine relation types, we choose \textit{xReact} and \textit{xWant} as generative relation representations. For example, given the input utterance \textit{``I'm not really that happy with you either''} and get the output \textit{xReact} and \textit{xWant} are \textit{``sad''} and \textit{``to be left alone''}, respectively.

\textbf{Relation Retrieval.} To retrieve relational knowledge, we apply SBERT \cite{reimers2019sentence} to compute the similar score between each utterance and each sentence in the ATOMIC dataset. After that, we select the phrases under the two relation types \textit{xReact} and \textit{xWant} of the most similar sentence as retrieved relation representations. In particular, the \textit{xReact} and \textit{xWant} phrases of the utterance \textit{``wait, it's- hey, stop... stop!''} are \textit{``frustrated''} and \textit{``to scold someone''}, respectively.

\textbf{Relation Encoding.} After obtaining the relation phrases, we put them into a combined template in order to receive the complete sentence $S_{rel}$. The combined template is formalized as:
\begin{equation}
    \begin{aligned}
        T_{xR}(u_i) = ``\textrm{The speaker feels }[xReact].'' \\
        T_{xW}(u_i) = ``\textrm{The speaker wants } [xWant].''
    \end{aligned}
\end{equation}
where $T_{.}(u_i)$ refers to the combined template of each relation type corresponding to the utterance $u_i$.

The complete sentences of generative and retrieved relation are separately fed to the BERT encoder to gain relation features. Finally, for each utterance $u_i$, we obtain four relation representations including $\textbf{c}_i^{xR}, \textbf{c}_i^{xW}, \textbf{s}_i^{xR}, \textbf{s}_i^{xW} \in \mathbb{R}^{l^R \times d}$, where $l^R$ denotes the length of the complete relation sentence.

\subsection{Textual Enhancement Module}
To take advantage of commonsense knowledge, we employ a Textual Enhancement Module (TEM) which integrates the relation features into textual features to enrich textual representations.

\textbf{Dual Perspective Learning.} We apply a dual perspective learning mechanism to perform relation fusion from two different views: generative and retrieved knowledge. First, we calculate learnable weight through a linear layer for each relation type. The formula is defined as follows:
\begin{equation}
    \begin{aligned}
        \alpha_{i}^{xR} = \textrm{SoftMax(}f_L\textrm{([}\textbf{h}_{i}^{T},\textbf{c}_i^{xR},\textbf{s}_i^{xR}\textrm{]))}\\
        \alpha_{i}^{xW} = \textrm{SoftMax(}f_L\textrm{([}\textbf{h}_{i}^{T},\textbf{c}_i^{xW},\textbf{s}_i^{xW}\textrm{]))}
    \end{aligned}
\end{equation}
where $\alpha_{i}^{xR}$, $\alpha_{i}^{xW}$ is the learnable weight corresponding to $xReact$ and $xWant$ relation, and $f_L$ denotes the linear layer.

Next, the relation fusion features are computed as follows:
\begin{equation}
    \begin{aligned}
        \textbf{h}_i^{xR} = \alpha_{i}^{xR} \cdot \textbf{c}_i^{xR} + (1-\alpha_{i}^{xR}) \cdot \textbf{s}_i^{xR}\\
        \textbf{h}_i^{xW} = \alpha_{i}^{xW} \cdot \textbf{c}_i^{xW} + (1-\alpha_{i}^{xW}) \cdot \textbf{s}_i^{xW}
    \end{aligned}
\end{equation}
where $\textbf{h}_i^{xR}, \textbf{h}_i^{xW} \in \mathbb{R}^{l^R \times d}$. 

\textbf{Textual Enhancing Fusion.} After obtaining the relation fusion features, we integrate them into the text feature by learning a trainable weight and tuning a hyper-parameter fused relation. For details, the formula is described as follows:
\begin{equation}
    \begin{aligned}
        \textbf{z}_i^{xR} = \textbf{h}_i^T + \mathbb{W}\textbf{h}_i^{xR}\\
        \textbf{z}_i^{xW} = \textbf{h}_i^T + \mathbb{W}\textbf{h}_i^{xW}
    \end{aligned}
\end{equation}
\begin{equation}
    \textbf{z}_i^{T} = \gamma \cdot \textbf{z}_i^{xR} + (1-\gamma) \cdot \textbf{z}_i^{xW}
    \label{eq:text-enhance}
\end{equation}
where $\textbf{z}_i^{T} \in \mathbb{R}^{l^S \times d}$ is the text-enhanced feature of utterance $u_i$, $\mathbb{W}$ denotes the trained weight, and $\gamma$ refers to the hyper-parameter.

\subsection{Multimodal Alignment Fusion}
Because of the independent learning of three modalities, we adopt a Multimodal Alignment Fusion (MAF) to align contextual information captured from separated modalities and fuse them to obtain the multimodal knowledge-enhanced representation of utterances.

First, to align the vision and acoustic feature with the text-enhanced feature, we apply the Connectionist Temporal Classification (CTC) \cite{graves2006connectionist} module:
\begin{equation}
    \textbf{z}_i^{T}, \textbf{z}_i^{V}, \textbf{z}_i^{A} = \textrm{CTC (} \textbf{z}_i^{T}, \textbf{h}_i^{V}, \textbf{h}_i^{A} \textrm{)}
\end{equation}
where $\textbf{z}_i^{T} \in \mathbb{R}^{l^S \times d}$, $\textbf{z}_i^{V} \in \mathbb{R}^{l^V \times d}$, $\textbf{z}_i^{A} \in \mathbb{R}^{l^A \times d}$ refer to the aligned features under each modality, and CTC is a module that consists of a LSTM block and a SoftMax function.

Subsequently, we concatenate the text-enhanced feature with visual and acoustic features. These concatenated features are then used to compute two filtering gates, which selectively emphasize relevant information within the visual and acoustic modalities, conditioned by the textual feature. The formulation is as follows:
\begin{equation}
    \begin{aligned}
        \textbf{g}_i^V = \textrm{ReLU ( } f_{VT} \textrm{( [ } \textbf{z}_i^V \mathbin\Vert \textbf{z}_i^T \textrm{ ] ) )}\\
        \textbf{g}_i^A = \textrm{ReLU ( } f_{AT} \textrm{( [ } \textbf{z}_i^A \mathbin\Vert \textbf{z}_i^T \textrm{ ] ) )}
    \end{aligned}
\end{equation}
where $\textbf{g}_i^V, \textbf{g}_i^A$ are two weighted gates related to the visual and acoustic features, ReLU is an activation function, $f_{*}$ denotes a linear layer and $\mathbin\Vert$ is notated for concatenating. 

Then, we produce the non-verbal feature by fusing the visual and acoustic features through two gates:
\begin{equation}
    \textbf{h}_i = \textbf{g}_i^V \cdot f_V \textrm{( } \textbf{z}_i^V \textrm{ ) } + \textbf{g}_i^A \cdot f_A \textrm{( } \textbf{z}_i^A \textrm{ ) }
\end{equation}
where $\textbf{h}_i \in \mathbb{R}^{l \times d}$, $l$ denotes the length of non-verbal token embeddings and $f_{*}$ is a linear layer.

Finally, we compute a fused weight $\beta$ between the text-enhanced feature and the non-verbal feature and then utilize it to create the multimodal feature $\bar{\textbf{z}} \in \mathbb{R}^{l \times d}$:
\begin{equation}
    \beta = min (\frac{\left\| \textbf{z}_i^T \right\|_2}{\left\| \textbf{h}_i \right\|_2}\varepsilon, 1)
\end{equation}
\begin{equation}
    \bar{\textbf{z}_i} = f(\textbf{z}_i^T + \beta\textbf{h}_i)
\end{equation}
where $\left\| . \right\|_2$ refers to $L_2$ normalization, $\varepsilon$ is a hyper-parameter, and $f$ denotes a normalized block including a layer normalization and dropout layer.

\subsection{Prediction and Loss Function}
\textbf{Prediction.} The output of the MAF module $\bar{\textbf{z}}$ is put through a Classifier to obtain the intent probability distribution. For details, the Classifier contains a pooling layer, a dropout layer, and the last one is a linear layer. The equation is described below:
\begin{equation}
    \hat{\textbf{y}_i} = f_c\textrm{ ( Dropout( Pooler( }\bar{\textbf{z}_i}\textrm{ ) ) )}
\end{equation}
where $\hat{\textbf{y}_i} \in \mathbb{R}^N$ denotes the predicted output, $N$ is the number of intent classes, and $f_c$ is a linear layer.

\textbf{Loss Function.} During the training phase, we apply a standard cross-entropy loss to optimize the performance of our model:
\begin{equation}
    \mathcal{L} = - \frac{1}{B} \sum_{i=1}^{B} \log \frac{\exp(\hat{\textbf{y}_i})}{\sum_{j=1}^{N}\exp(\hat{\textbf{y}_j})}
\end{equation}
where $B$ is the batch size, and $\hat{\textbf{y}_i}$ denotes the predicted label of $i^{th}$ sample.

\section{Experiments}

\begin{table*}
    \centering
    \caption{Multimodal intent recognition results on the MIntRec dataset. ``Twenty-class'' and ``Binary-class'' denote the multi-class and binary classification. The best performances are highlighted in \textbf{bold}, and the \underline{underline} refers to the second-best ones. Results with * are obtained by reimplemented, while others are taken from the corresponding published paper.}
    \begin{adjustbox}{width=\columnwidth*2,center}
        \begin{tabular}{l|cccc|cccc}
        \toprule
        \multicolumn{1}{c|}{\multirow{2}{*}{Methods}} & \multicolumn{4}{c|}{Twenty-class} & \multicolumn{4}{c}{Binary-class} \\ 
        \multicolumn{1}{c|}{} & \multicolumn{1}{c}{ACC (\%)} & \multicolumn{1}{c}{F1 (\%)} & \multicolumn{1}{c}{PREC (\%)} & \multicolumn{1}{c|}{REC (\%)} & \multicolumn{1}{c}{ACC (\%)} & \multicolumn{1}{c}{F1 (\%)} & \multicolumn{1}{c}{PREC (\%)} & \multicolumn{1}{c}{REC (\%)} \\ 
        \midrule
        Text Classifier & 70.88 & 67.40 & 68.07 & 67.44 & 88.09 & 87.96 & 87.95 & 88.09 \\ 
        MAG-BERT & \textbf{72.65} & 68.64 & 69.08 & \underline{69.28} & 89.24 & 89.10 & 89.10 & 89.13 \\ 
        MulT & \underline{72.52} & 69.25 & 70.25 & 69.24 & 89.19 & 89.01 & 89.02 & \underline{89.18} \\ 
        MISA & 72.29 & \underline{69.32} & \textbf{70.85} & 69.24 & 89.21 & 89.06 & 89.12 & 89.06 \\ 
        SDIF-DA* & 71.01 & 67.77 & 68.75 & 67.7 & 88.76 & 88.65 & 88.56 & 88.77 \\ 
        TCL-MAP* & 71.46 & 68.02 & 67.84 & 69.23 & \underline{89.44} & \underline{89.26} & \underline{89.44} & 89.11 \\ 
        \midrule
        \textbf{TECO (Ours)} & 72.36 & \textbf{69.96} & \underline{70.49} & \textbf{69.92} & \textbf{89.66} & \textbf{89.54} & \textbf{89.5} & \textbf{89.58} \\ 
        \bottomrule
        \end{tabular}
    \end{adjustbox}
    \label{tab:main_results}
\end{table*}
\subsection{Experimental Settings}
\textbf{Dataset.} We conduct experiments on MIntRec \cite{zhang2022mintrec} dataset which is a fine-grained dataset for multimodal intent recognition. This dataset comprises $2,224$ high-quality samples with three modalities: text, vision, and acoustic across twenty intent categories. The dataset is divided into a training set of $1,334$ samples, a validation set of $445$ samples, and a test set of $445$ samples.

\textbf{Implementation Details.} For the implementation of our proposed method, we set the training batch size is $16$, while the validation and test batch sizes are both $8$. The number of epochs for training is set to $100,$ and we apply early stopping for $8$ epochs. To optimize the parameters, we employ an AdamW \cite{loshchilov2017decoupled} optimizer with linear warm-up and a weight decay of $1e-2$ for parameter tuning. The initial learning rate is set to $2e-5$ and the hyper-parameter fused relation $\gamma$ is chosen from $[0.05:0.95]$. As sequence lengths of the segments in each modality and relation sentence need to be fixed, we use zero-padding for shorter sequences. $l^S, l^V, l^A, l^R$ are $30, 230, 480,$ and $30$, respectively.

\textbf{Evaluation Metrics.} We use four metrics to evaluate our model performance: accuracy (ACC), F1-score (F1), precision (PREC), and recall (REC). The macro score over all classes for the last three metrics is reported. The higher values indicate improved performance of all metrics.

\subsection{Baselines}
We compare our framework with several comparative baseline methods: 
\begin{itemize}
    \item \textbf{Text Classifier} \cite{zhang2022mintrec} is a classifier with text-only modality that uses the first special token $[CLS]$ from the last hidden layer of the BERT pre-trained model as the sentence representation.
    \item \textbf{MAG-BERT} \cite{rahman2020integrating} integrated the two non-verbal features including video and acoustic features into the lexical one by applying a Multimodal Adaptation Gate (MAG) module attached to the BERT structure.
    \item \textbf{MulT} \cite{tsai2019multimodal} stands for the Multimodal Transformer, an end-to-end model that extends the standard Transformer network \cite{vaswani2017attention} to learn representations directly from unaligned multimodal streams.
    \item \textbf{MISA} \cite{hazarika2020misa} projected each modality to two distinct subspaces. The first one learns their commonalities and reduces the modality gap, while the other is private to each modality and captures their characteristic features. These representations provide a holistic view of the multimodal data.
    \item \textbf{SDIF-DA} \cite{huang2024sdif} is a Shallow-to-Deep Interaction Framework with Data Augmentation that effectively fuses different modalities’ features and alleviates the data scarcity problem by utilizing the shallow interaction and the deep one.
    \item \textbf{TCL-MAP} \cite{zhou2024token} proposed a modality-aware prompting module (MAP) to align and fuse features from text, video, and audio modalities with the token-level contrastive learning framework (TCL).
\end{itemize}
\begin{table*}[t!]
    \centering
    \caption{Ablation experiments of several modules within our model on both multi-class and binary classification stages.}
    \begin{adjustbox}{width=\columnwidth*2,center}
        \begin{tabular}{l|cccc|cccc}
        \toprule
        \multicolumn{1}{c|}{\multirow{2}{*}{Methods}} & \multicolumn{4}{c|}{Twenty-class} & \multicolumn{4}{c}{Binary-class} \\   
        \multicolumn{1}{c|}{} & \multicolumn{1}{c}{ACC (\%)} & \multicolumn{1}{c}{F1 (\%)} & \multicolumn{1}{c}{PREC (\%)} & REC (\%) & \multicolumn{1}{c}{ACC (\%)} & \multicolumn{1}{c}{F1 (\%)} & \multicolumn{1}{c}{PREC (\%)} & \multicolumn{1}{c}{REC (\%)} \\ 
        \midrule
        \textbf{TECO (Ours)} & \textbf{72.36} & \textbf{69.96} & \textbf{70.49} & \textbf{69.92} & \textbf{89.66} & \textbf{89.54} & \textbf{89.5} & \textbf{89.58}\\ 
        \midrule
        $w_{TV}$ & 70.79 & 66.05 & 66.35 & 66.77 & 88.54 & 88.35 & 88.48 & 88.26 \\ 
        $w_{TA}$ & 70.34 & 66.91 & 67.49 & 67.04 & 88.99 & 88.85 & 88.83 & 88.87 \\ 
        $w_{VA}$ & 16.85 & 3.16 & 2.46 & 6.66 & 52.36 & 48.28 & 49.75 & 49.79 \\ 
        $w/o_{TEM}$ & 70.34 & 64.4 & 64.43 & 65.03 & 88.54 & 88.45 & 88.33 & 88.67 \\ 
        $w/o_{MAF}$ & 71.91 & 68.19 & 68.67 & 68.45 & 87.42 & 87.33 & 87.22 & 87.61 \\ 
        $w/o_{dual}$ & 69.44 & 65.68 & 66.07 & 65.83 & 87.19 & 87.04 & 86.99 & 87.1 \\ 
        \bottomrule
        \end{tabular}
    \end{adjustbox}
    \label{tab:ablation}
\end{table*}
\subsection{Results}
Table \ref{tab:main_results} describes the results conducted on the intent recognition tasks. Overall, our approach gains significant performances compared to the baselines on the two tasks: binary classification and multi-class classification. Especially, in the binary classification stage, our method outperforms the others across all four metrics. Compared to the second-best methods, the considerable enhancements of $0.25\%$ on accuracy, $0.31\%$ on macro F1-score, $0.67\%$ on precision, and $0.53\%$ on recall indicate the efficiency of our model to leverage multimodal information for understanding real-world context. In the remaining task, our method achieves notable improvements on two metrics macro F1-score and recall, and also gains the second-best result on precision. This observation illustrates the capability of our proposed model in recognizing speakers' intents within a dialog act. 

\subsection{Ablation Study}
\subsubsection{Contribution Analysis of Model Components}
To further analyze the contributions of each component to overall performance, we conduct a set of ablation studies including setting model with \textit{(1)} text and video information ($w_{TV}$), \textit{(2)}
text and audio features ($w_{TA}$), and \textit{(3)} video combined with audio representation ($w_{VA}$); removing \textit{(4)} the Text Enhancement Module ($w/o_{TEM}$), \textit{(5)} the Multimodal Alignment Fusion module ($w/o_{MAF}$), and \textit{(6)} the dual perspective learning by detaching SBERT component($w/o_{dual}$).

\textbf{The important role of the text representation.} We explore the role of modalities by removing one modality at a time in ablation studies \textit{(1), (2), (3)}. As shown in Table \ref{tab:ablation}, the accuracy of our methods decreased seriously when the contextual modality was removed. Particularly, similar drops in performance are not observed then other two modalities are removed, which indicates that textual information has a dominant effect. 

\textbf{The effect of dual perspective learning and textual enhancement module.} To explore whether the dual perspective learning, we conduct an experiment \textit{(6)} that removes retrieved relation from SBERT and remains generative relation extracted from COMET to enhance text representation without dual-view. We can observe that the TECO without dual perspective learning experiences a significant lessening of $4.2\%$ and $2.8\%$ in accuracy for multi-class and binary-class classification, respectively. In addition, we remove features obtained from both COMET and SBERT which is described in experiment \textit{(4)} to prove the necessary role of commonsense knowledge. We can observe that the final result witnessed a substantial decrease in most metrics indicating that our method is successful in strengthening verbal representation.

\textbf{MAF works productively in multimodal fusion operation.} In experiment \textit{(5)}, we assess the effectiveness of multimodal alignment fusion by discharging both two non-verbal features. As indicated by the results, the performance shows a reduction of more than $2\%$ across most metrics for multi-class. The same trend was witnessed in several metrics for binary classification. The experimental results illustrate that contextual modality plays a critical role in integrating and predicting user's intents.

\subsubsection{Hyper-parameter Analysis}
\begin{figure}[t!]
    % \centering
    \includegraphics[width=0.9\linewidth]{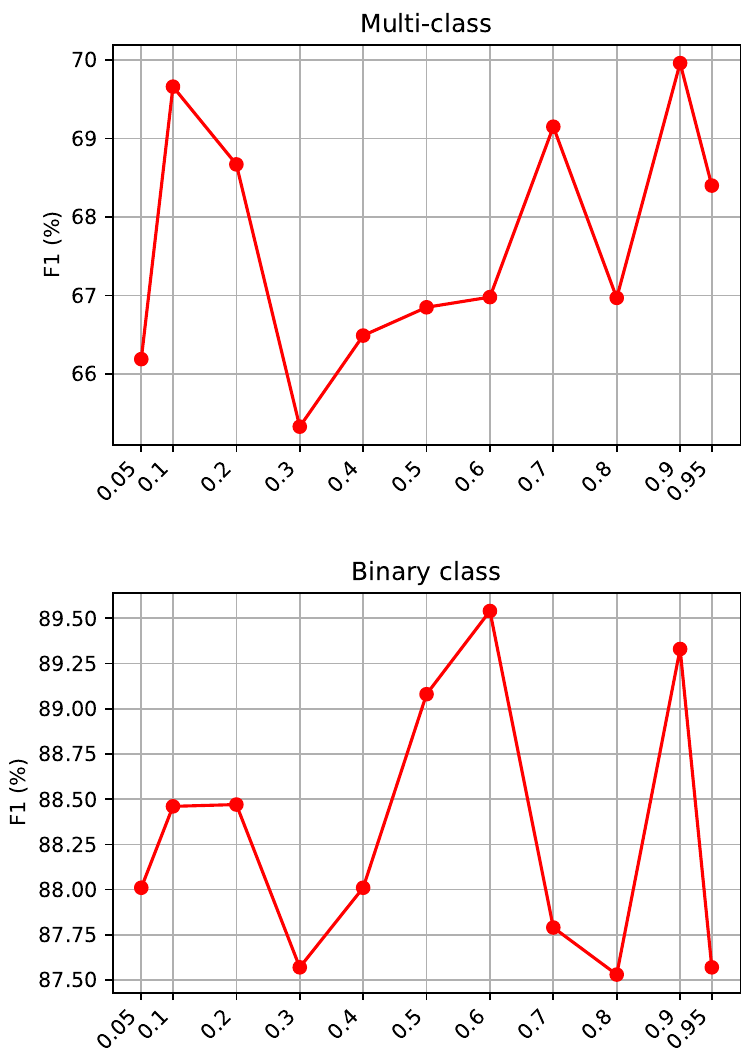}
    \caption{Model performance across different value of $\gamma$}
    \label{fig:gamma}
\end{figure}

To evaluate the influence of each relation type on our model's performance, we set up experiments by changing the hyperparameter $\gamma$ in Equation \ref{eq:text-enhance}. The results are recorded in Figure \ref{fig:gamma}, which the former is conducted for multi-class classification while the latter is for binary one. We find that macro F1-score is improved at $\gamma = 0.9$ and $\gamma = 0.6$ on muti-class and binary class, respectively. This indicates the relation $xReact$ having more effect on enhancing text representations and boosting the model capability of detecting intention than the relation $xWant$.

\subsection{Case Study}

\begin{table*}[t!]
    \centering
    \caption{The illustration of case studies, where the text with green color indicates the correct prediction, while the other is the incorrect one.}
    \begin{adjustbox}{width=\columnwidth*2,center}
        \begin{tabular}{m{3.5cm}|m{4cm}|m{4cm}|m{1.25cm}m{1.25cm}|m{1.25cm}m{1.25cm}|m{1.25cm}m{1.25cm}}
        \toprule
        \multirow{2}{*}{\textbf{Text}} & \multirow{2}{*}{\textbf{Video}} & \multirow{2}{*}{\textbf{Audio}} & \multicolumn{2}{c}{\textit{xReact}} & \multicolumn{2}{|c}{\textit{xWant}} & \multicolumn{2}{|c}{Intent} \\
        \multicolumn{1}{c|}{} & & & \textbf{COMET} & \textbf{SBERT} & \textbf{COMET} & \textbf{SBERT} & \textbf{Label} & \textbf{Predicted}  \\
        \midrule
         ``Yeah, those babies look great.'' & \includegraphics[width=1\linewidth]{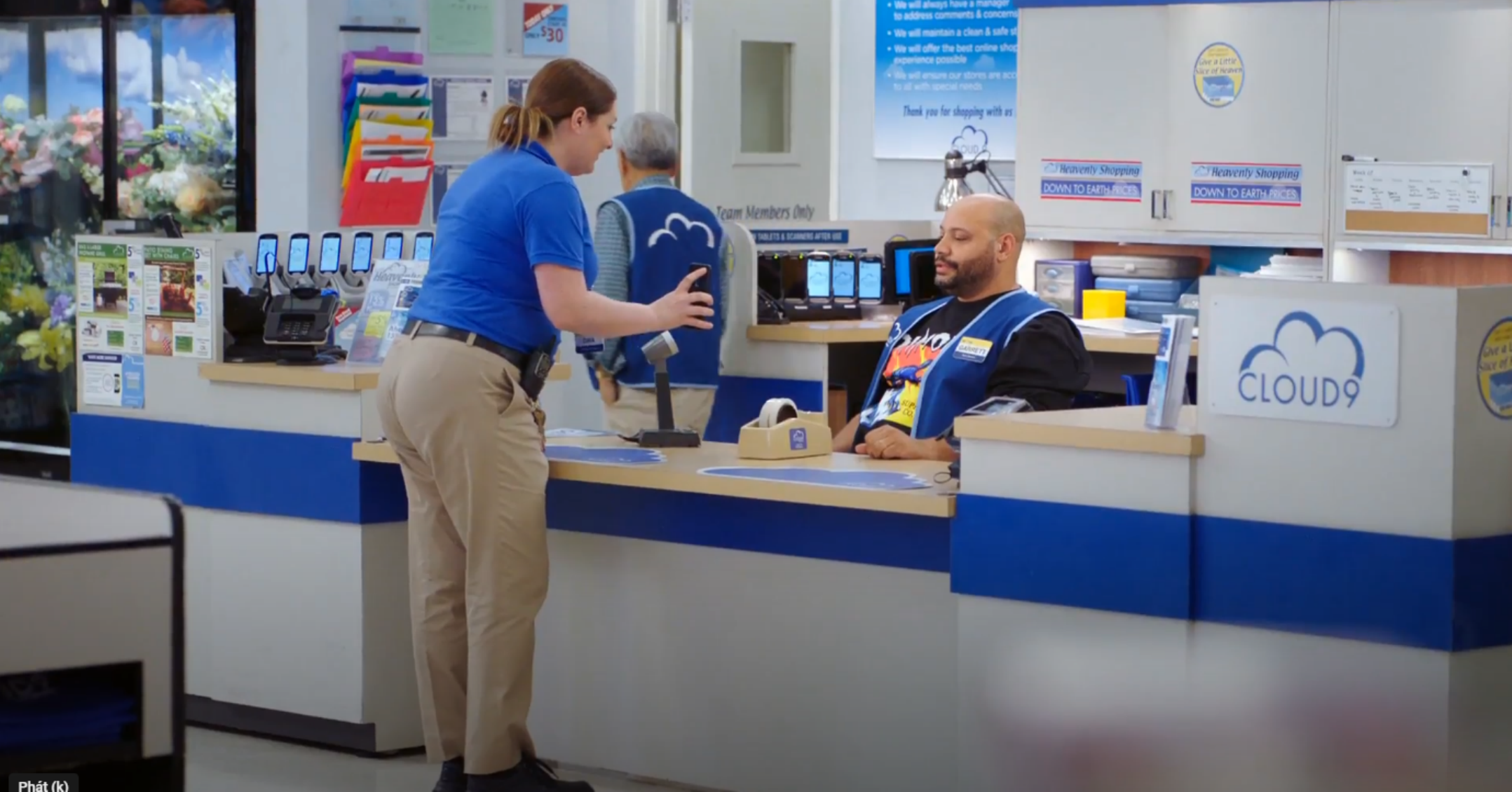} & \includegraphics[width=1\linewidth]{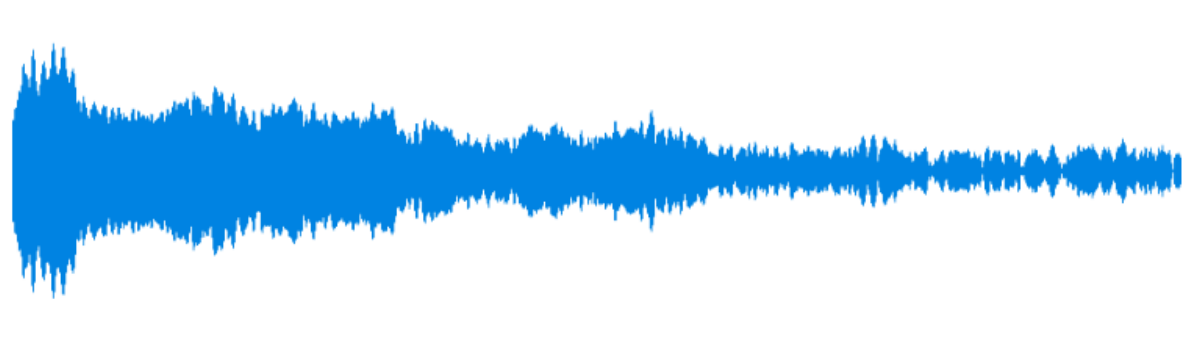}& happy & very happy & to have a good time & smile at the baby & Praise & \textbf{Praise} \\
         \midrule
         ``And unfortunately, it is supposed to rain.'' & \includegraphics[width=1\linewidth]{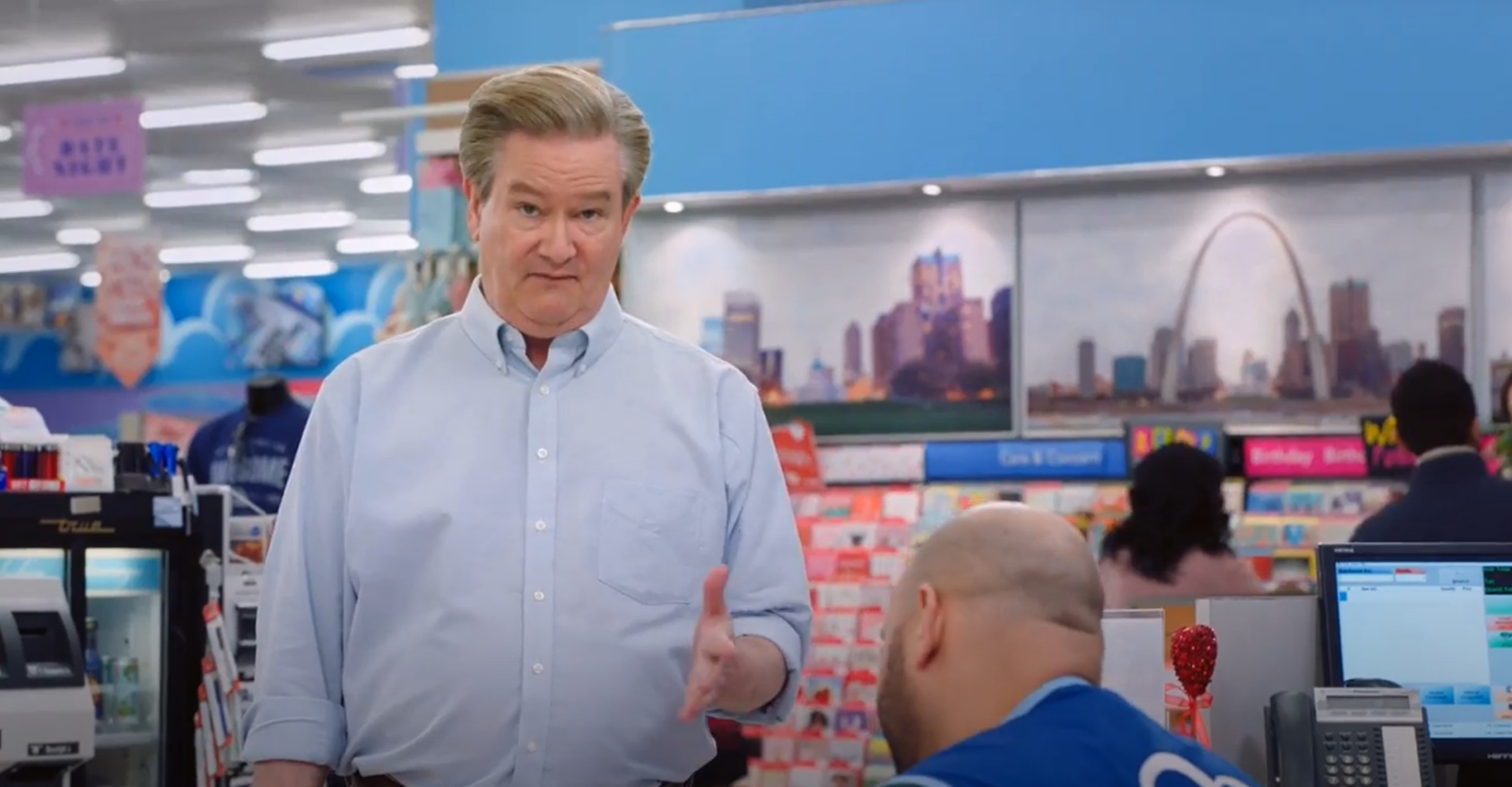} & \includegraphics[width=1\linewidth]{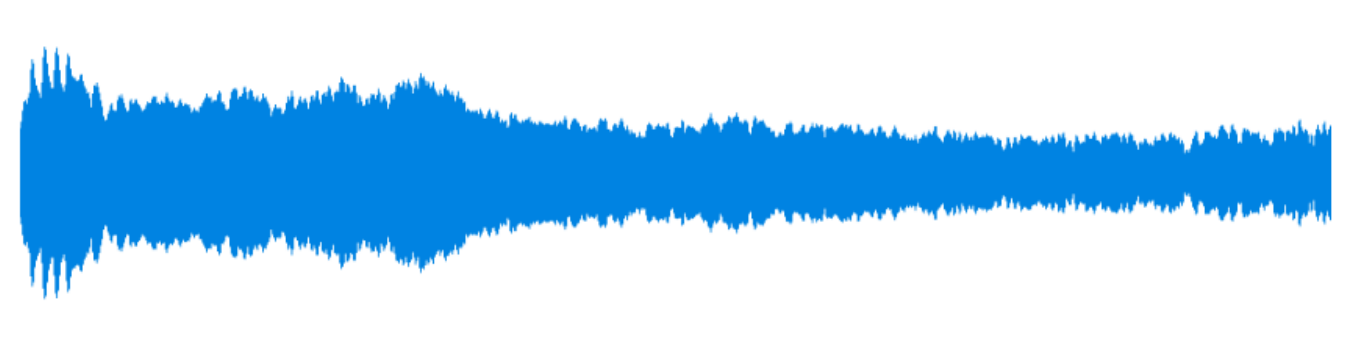} & sad & very worry & to get a umbrella & to stay dry & Complain & \textbf{Complain}\\
         \midrule
         ``So thank you all so much for my gifts.'' & \includegraphics[width=1\linewidth]{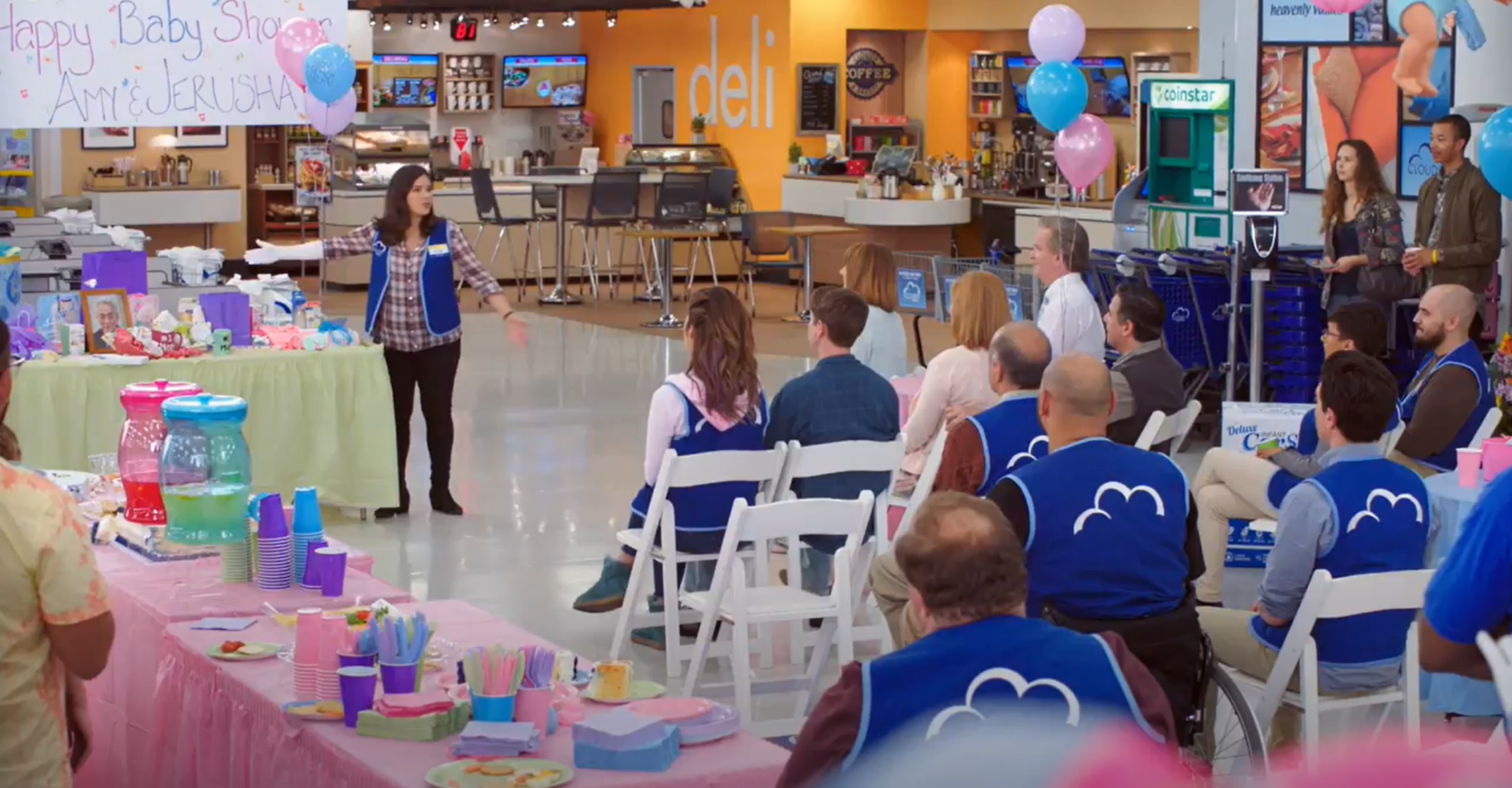} & \includegraphics[width=1\linewidth]{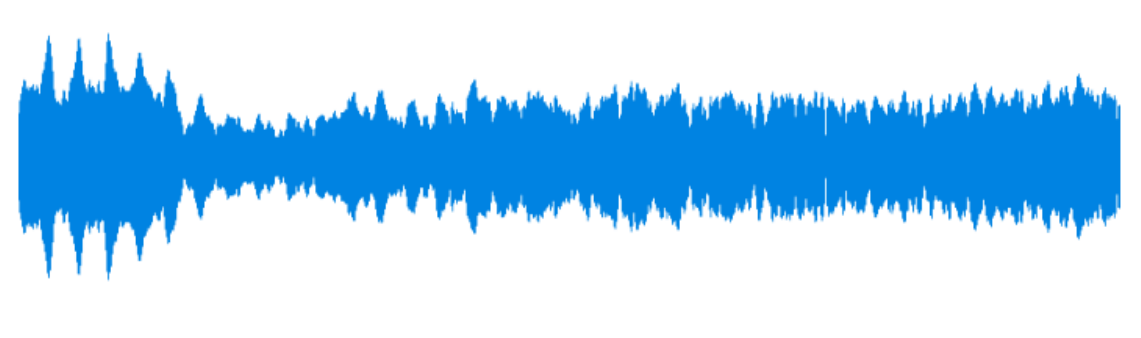} & happy & happy & to show appreciation & to accept the givings & Thank & \textbf{Thank} \\
         \midrule
         ``Stop, please.'' & \includegraphics[width=1\linewidth]{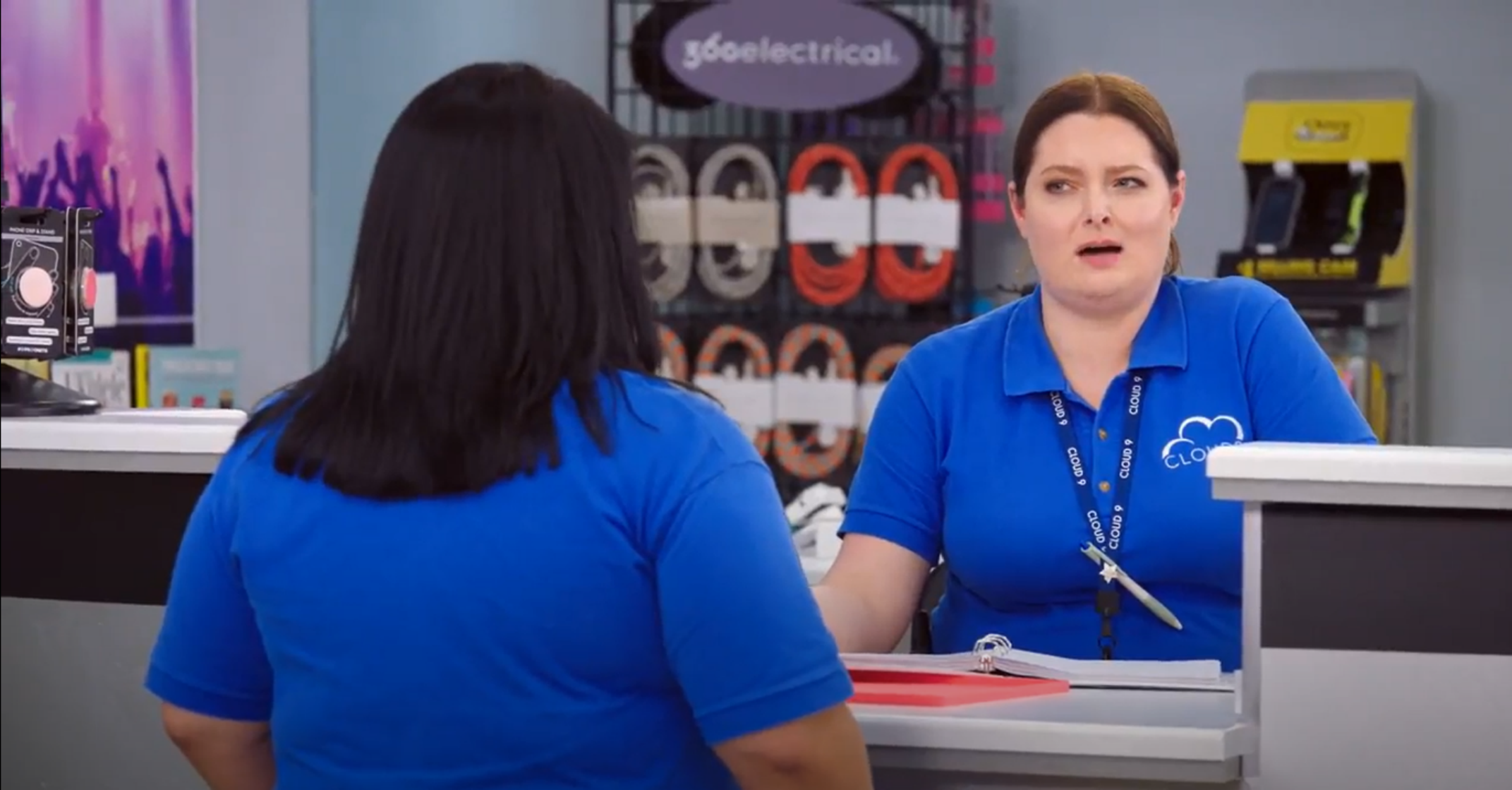} & \includegraphics[width=1\linewidth]{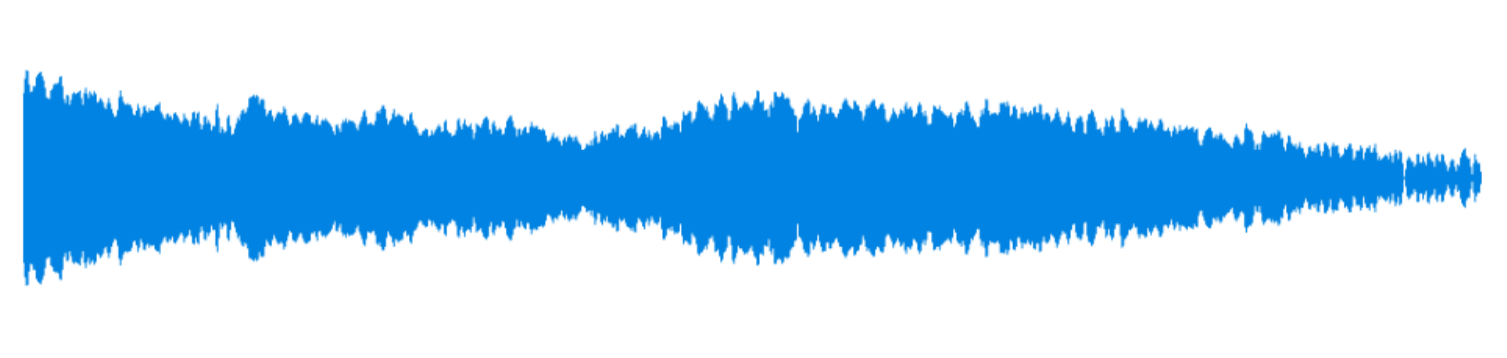} & happy & scared & to be a good friend & to get away & Prevent & \textbf{Oppose}\\ 
         \midrule
         ``Hey, we have a problem.'' & \includegraphics[width=1\linewidth]{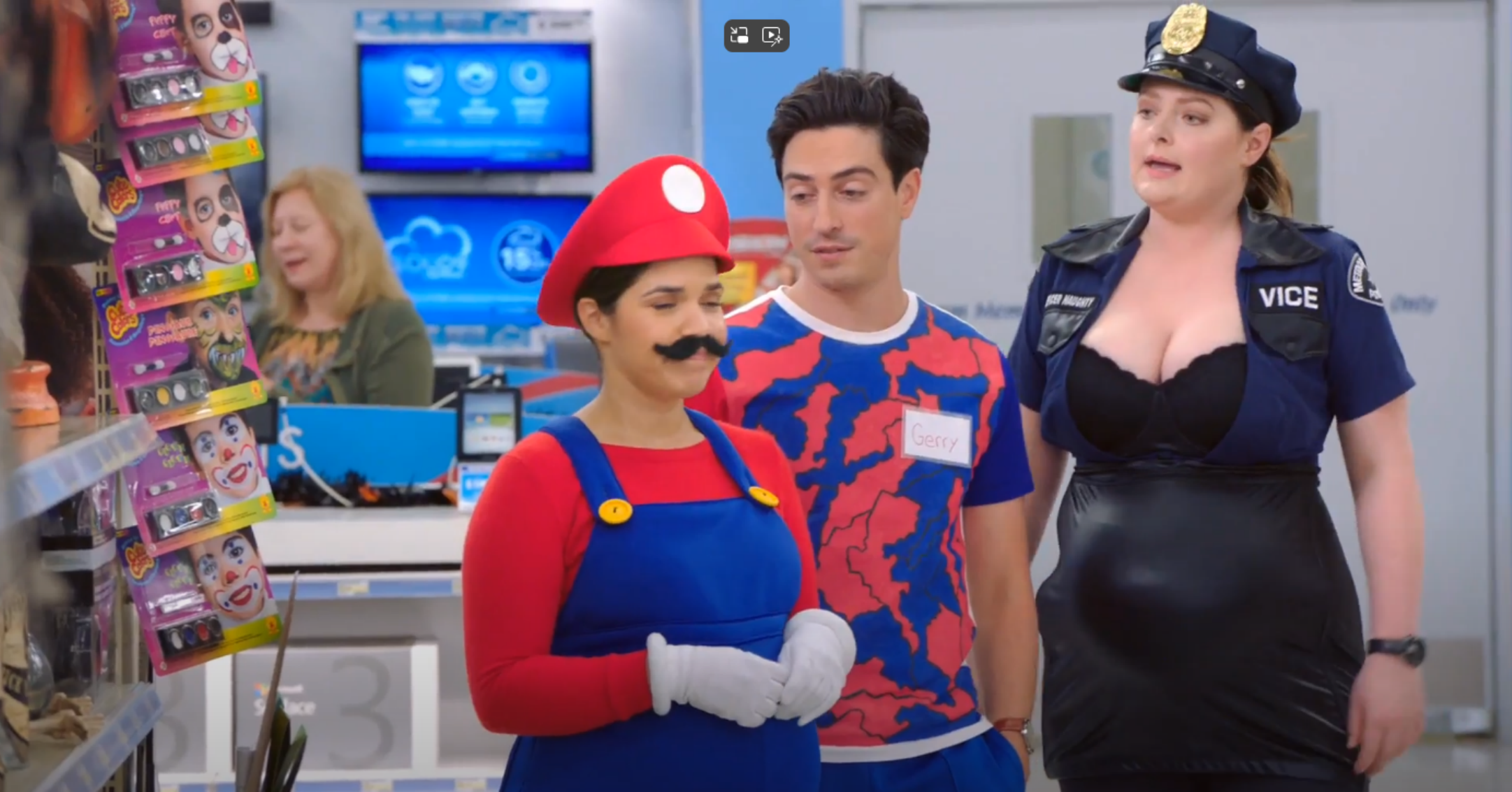} & \includegraphics[width=1\linewidth]{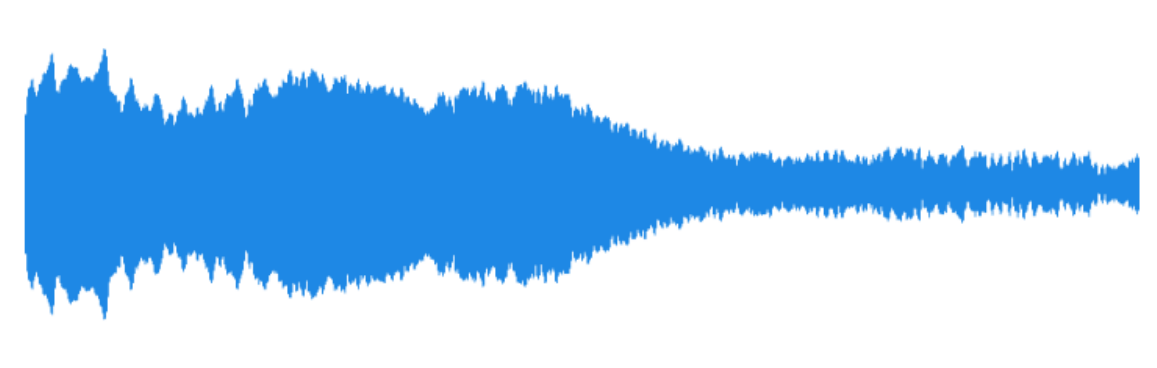} & worried & curious & to solve the problem & to make adjustments & Inform & \textbf{Ask for help}\\
         \bottomrule
        \end{tabular}
    \end{adjustbox}
    \label{tab:case}
\end{table*}

To demonstrate the association and impact of the two relations $xReact$ and $xWant$ derived from generative and retrieved knowledge extractor, we write down several samples in Table \ref{tab:case}. The first three examples show the relevance between relation and label intent, which make the donation of producing the correct prediction. Especially, $xReact$ tends to express feelings related to intention, while $xWant$ is able to generalize the meanings of the sentence. Our COKE module can generate relations more precisely with ``expressing emotions'' intents such as \textit{Praise, Complain, Thank} than ``achieving goals'' such as \textit{Inform}, \textit{Prevent}. In addition, obtaining relations from sentences with clear emotional words is more exact than from those that are brief and ambiguous.

\section{Conclusion}
In this work, we introduce a Text Enhancement associated with Commonsense Knowledge Extractor (TECO) for multimodal intent recognition. Our model enriches text information by integrating relation information extracted from a commonsense knowledge graph. Thanks to the strength of commonsense knowledge, the implicit contexts of input utterances are explored and utilized to enhance verbal representations. In addition, both visual and acoustic representations are aligned with textual ones to obtain consistent information and then fused together to gain meaningful and rich multimodal features. To evaluate our method's performance, we conducted several experiments and ablation studies on the MIntRec dataset and achieved remarkable results.

% \section*{Acknowledgments}

\bibliography{references}

\end{document}